\documentclass[runningheads]{llncs}
\usepackage{graphicx}

\usepackage[export]{adjustbox}
\usepackage{booktabs}
\usepackage{amsmath}
\usepackage{hyperref}
\usepackage{pifont}
\usepackage{lipsum,booktabs}
\usepackage{capt-of}
\usepackage{tabu}

\newcommand{\cmark}{\ding{51}}%
\newcommand{\xmark}{\ding{55}}%
\begin{document}
%


\title{Reading Between the Lanes: Text VideoQA on the Road}
%
%

\author{George Tom\inst{1} \orcidID{0009-0002-7343-1680} \and
Minesh Mathew\inst{1} \orcidID{0000-0002-0809-2590} \and
Sergi Garcia-Bordils\inst{2,3} \orcidID{0000-0002-4222-8367} \and
Dimosthenis Karatzas\inst{2} \orcidID{0000-0001-8762-4454} \and
C.V. Jawahar\inst{1}\orcidID{0000-0001-6767-7057}}  
\authorrunning{G. Tom et al.}


\institute{
Center for Visual Information Technology (CVIT), IIIT Hyderabad, India
\email{\{george.tom,minesh.mathew\}@research.iiit.ac.in, jawahar@iiit.ac.in}
\and
Computer Vision Center (CVC), UAB, Spain
\email{\{sergi.garcia,dimos\}@cvc.uab.cat}
\and
AllRead Machine Learning Technologies
} 

\maketitle              
\begin{abstract}
Text and signs around roads provide crucial information for drivers, vital for safe navigation and situational awareness.
Scene text recognition in motion is a challenging problem, while textual cues typically appear for a short time span, and early detection at a distance is necessary. Systems that exploit such information to assist the driver should not only extract and incorporate visual and textual cues from the video stream but also reason over time.
To address this issue, we introduce RoadTextVQA, a new dataset for the task of video question answering (VideoQA) in the context of driver assistance. RoadTextVQA consists of $3,222$ driving videos collected from multiple countries, annotated with $10,500$ questions, all based on text or road signs present in the driving videos. We assess the performance of state-of-the-art video question answering models on our RoadTextVQA dataset, highlighting the significant potential for improvement in this domain and the usefulness of the dataset in advancing research on in-vehicle support systems and text-aware multimodal question answering. The dataset is available at \href{http://cvit.iiit.ac.in/research/projects/cvit-projects/roadtextvqa}{http://cvit.iiit.ac.in/research/projects/cvit-projects/roadtextvqa}

\end{abstract}
\section{Introduction}
\label{sec:intro}

\begin{figure}[t]
    \centering
    \includegraphics[width=\textwidth]{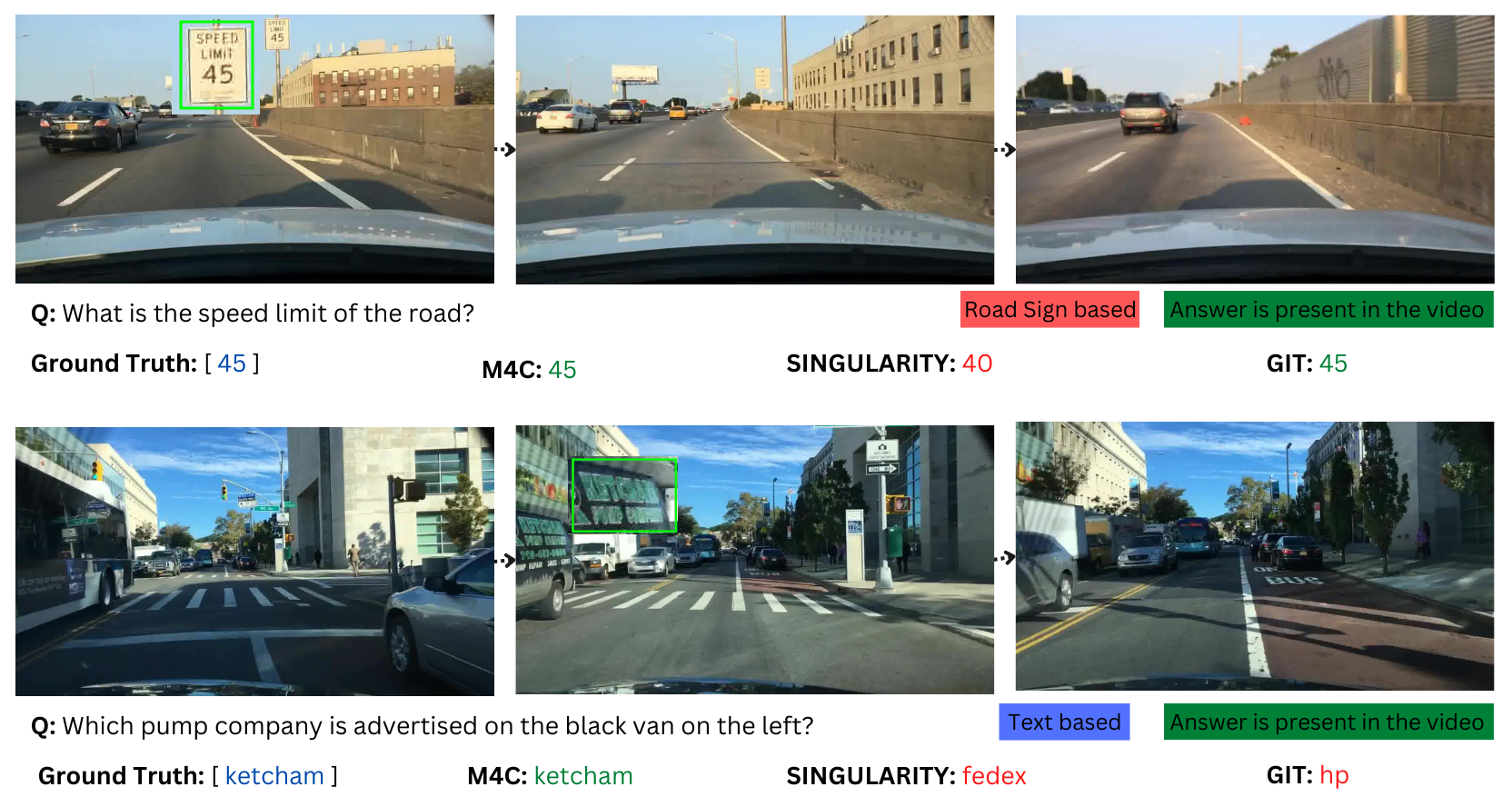}
    \caption{Examples from our RoadTextVQA dataset. The question in the first clip is based on the speed limit road sign, so it is classified as a ``road sign based" question. Meanwhile, the question for the clip in the second row draws information from the text on the van, making it a ``text based" question. The ground truth answers and the baseline predictions are also presented.}\label{main_label} \label{roadtext_main}
\end{figure}

In this work, we propose a new dataset for Visual Question Answering (VQA) on driving videos, with a focus on questions that require reading text seen on the roads and understanding road signs. Text and road signs provide important information to the driver or a driver assistance system and help to make informed decisions about their route, including how to reach their destination safely and efficiently. Text on roads can also provide directions, such as turn-by-turn directions or the distance to a destination. Road signs can indicate the location of exits, rest stops, and potential hazards, such as road construction or detours. Reading text and understanding road signs is also important for following traffic laws and regulations. Speed limit signs, yield signs, and stop signs provide important information that drivers must follow to ensure their own safety and the safety of others on the road.

VQA is often dubbed as the Turing test for image/video understanding. The early datasets for VQA on images and videos~\cite{vqav1,tapaswi2016movieqa,xu2017video} largely ignored the need for reading and comprehending text on images and videos, and questions were mostly focus on the visual aspects of the given image or video. For example, questions focused on the type, attributes and names of objects, things or people. However, the text is ubiquitous in outdoor scenes, and this is evident from the fact that nearly $50\%$ of the images in the MS-COCO dataset have text in them~\cite{cocotext}.
Realizing the importance of reading text in understanding visual scenes, two datasets---Scene text VQA~\cite{Biten_2019_ICCV} and Text VQA~\cite{textvqa} were introduced that focus exclusively on VQA involving scene text in natural images. 
Two recent works called NewsVideoQA\cite{newsvideoqa}, and M4-ViteVQA\cite{singularity} extend text-based VQA works to videos by proposing VQA tasks   that exclusively focus on question-answers that require systems to read the text in the videos.

Similar to these works that focus on text VQA on videos, our work proposes a new dataset where all the questions need to be answered by watching driving videos and reading the text in them. However, in contrast to NewsVideoQA which contains news videos where question-answer pairs are based on video text (born-digital embedded text) appearing on news tickers and headlines, the text in videos in our dataset are scene text. The text in the road  or driving videos  are subjected to blur, poor contrast, lighting conditions and distortions. Text while driving goes by fast and tends to be heavily occluded. Often, multiple frames needs to be combined to reconstruct the full text, or a good frame with readable text needs to be retrieved. These difficulties made researchers focus on road-text recognition exclusively, and there have been works that focus exclusively on the detection, recognition and tracking of road text videos~\cite{roadtext1k,RoadText-3k}. On the other hand M4-ViteVQA contains varied type of videos such as sports videos, outdoor videos and movie clips. A  subset of these videos are driving videos. In contrast, our dataset is exclusively for VQA on driving videos and contains at least three times more questions than in the driving subset of M4-ViteQA. Additionally, questions in our dataset require both reading road text and understanding road signs, while M4-ViteVQA's focus is purely on text-based VQA.\\

Specifically our contributions are the following:

\begin{itemize}
    \item We introduce  the first large scale dataset for road text and road sign VQA containing 10K+ questions and 3K+ videos.

    \item We provide a thorough analysis of the dataset and present detailed statistics of videos, questions and answers. We also establish heuristic baselines and upper bounds that help to estimate the difficulty of the problem.

    \item We evaluate an existing popular VQA model and two  SoTA VideoQA models on our dataset and demonstrate that these models fail to perform well on the new dataset since they are not designed to read and reason about text and road signs.
\end{itemize}

\section{Related Work}
\label{sec:related_works}

\subsection{VideoQA}
\label{sec:related_videoVQA}
In video question answering(VideoQA), the goal is to answer the question in the context of the video. Earlier approaches to VideoQA use LSTM to encode the question and videos\cite{zhao2018multi,li2019beyond,kim2020modality,wang2018movie}. 
Several datasets have been created in recent years to assist research in the field of video question answering (VideoQA). Large datasets such as MSRVTT-QA\cite{xu2017video} contain synthetic generated questions and answers where the questions require only an understanding of the visual scenes. MOVIE-QA\cite{tapaswi2016movieqa} and TVQA\cite{lei2018tvqa} are based on scenes in movies and TV shows. Castro et al.\cite{castro2022wild} introduced a dataset with videos from the outside world for video understanding through VideoQA and Video Evidence Selection for interpretability. MOVIE-QA\cite{tapaswi2016movieqa}, TVQA\cite{lei2018tvqa}, HowtoVQA69M\cite{yang2021justask} 
provide explicit text in the form of subtitles. Multiple-Choice datasets\cite{tapaswi2016movieqa,lei2018tvqa,xu2021sutd}  consist of a pre-defined set of options for answers. When compared to open-ended datasets, they can be considered limiting in the context of real-world applications. Synthetically generated datasets\cite{xu2017video,yu2019activityqa,castro2022wild} contain questions that are generated through processing video descriptions, narration and template questions. MSRVTT-QA\cite{xu2017video} exploits the video descriptions for QA creation. HowToVQA69M\cite{yang2021justask} uses cross-modal supervision and language models to generate question-answer pairs from narrated videos, whereas ActivityNetQA\cite{yu2019activityqa} uses template questions to generate the QA pairs. Xu et al. introduced the SUTD-TrafficQA\cite{xu2021sutd} dataset and the Eclipse model for testing systems' ability to reason over complex traffic scenarios. The SUTD-TrafficQA\cite{xu2021sutd} dataset contains multiple-choice questions that are based on different traffic events.  RoadTextVQA is an open-ended dataset that deals with questions related to the text information found in road videos or the signs posted along roads. Recent studies\cite{singularity,clipbert,hero,blip} on pretraining transformers on other vision and language tasks have shown excellent results for the VideoQA task. Lei et al. \cite{singularity}, in their study, uncovered the bias present in many video question-answering datasets, which only require information from a single frame to answer, and introduced new tasks aimed at training models to answer questions that necessitate the use of temporal information.

\begin{table*}[!t]
    \caption{Comparison of RoadTextVQA with existing video question answering datasets. ``Text-based" indicates whether the questions require an understanding of the text present in the videos to answer. ``Road-based" questions are datasets which are based on the driving domain. ``Synthetic questions" are questions that are not manually annotated and depend on automated methods for question-answer generation. Abbreviations used - OE: Open-ended questions, MC: Multiple choice questions.}
    \centering
    \begin{adjustbox}{width=1\textwidth}
    \begin{tabular}{@{}lcccccc@{}}
    \toprule
    \textbf{Dataset} & \textbf{  Text-based  } & \textbf{  Road-based  } & \textbf{  Synthetic Questions  } & \textbf{  \#Videos  } & \textbf{  \#Questions  } & \textbf{  QA type  } \\
    \midrule
    MovieQA\cite{tapaswi2016movieqa} & \xmark & \xmark & \xmark & 6.7K & 6.4K & MC \\ 
    MSRVTT-QA\cite{xu2017video} & \xmark & \xmark & \cmark & 10K  & 243.6K & OE \\
    Activitynet-QA\cite{yu2019activityqa} & \xmark & \xmark & \cmark & 5.8K  & 58K & OE \\
    TVQA\cite{lei2018tvqa} & \xmark & \xmark & \xmark  & 21.7K  & 152.5K & MC \\
    WildQA\cite{castro2022wild} & \xmark & \xmark & \xmark  & 0.4K  & 0.9K & OE \\
    HowtoVQA69M\cite{yang2021justask} & \xmark & \xmark & \cmark  & 69M  & 69MK & OE \\
    SUTD-TrafficQA\cite{xu2021sutd} & \xmark & \cmark & \xmark  & 10K & 62.5K & MC \\
    NewsVideoQA\cite{newsvideoqa} & \cmark & \xmark & \xmark  & 3K & 8.6K & OE \\
    M4-ViteVQA\cite{zhao2022towards} & \cmark & \xmark & \xmark & 7.6K  &  25.1K & OE \\
    \textbf{RoadTextVQA} & \cmark & \cmark & \xmark & 3.2K & 10.5K & OE \\
    \bottomrule
    \end{tabular}
    \end{adjustbox}
    \label{tab:roadtext1k_comparison}
\end{table*}

\subsection{VideoQA involving video text}
\label{sec:related_videotextvqa}
NewsVideoQA\cite{jahagirdar2023watching} and M4-ViteVQA\cite{zhao2022towards} are two recently introduced datasets that include videos with embedded born-digital text and scene text, respectively. Both datasets require an understanding of the text in videos to answer the questions. 
Embedded text, sometimes called video text in news videos, is 
often displayed with good contrast and in an easy-to-read style. 
Scene text in the RoadTextVQA dataset can be challenging to read due to the factors such as occlusion, blur, and perspective distortion. M4-ViteVQA contains videos from different domains, a few of them being shopping, driving, sports, movie and vlogs. The size of RoadTextVQA is more than three times the size driving subset of M4-ViteVQA. Additionally, a subset of questions in RoadTextVQA also requires domain knowledge to answer questions related to road signs. Few recent works\cite{wang2022git,pali} on vision and language transformers have shown to work well with text-based VQA tasks. Kil et al.\cite{prestu} introduced PreSTU, a pretraining method that improves text recognition and connects the recognized text with the rest of the image. GIT(GenerativeImage2Text)\cite{wang2022git} is a transformer-based model for vision and language tasks with a simple architecture that does not depend on external OCR or object detectors. 

\subsection{Scene Text VQA}
Our work, which focuses on VQA requiring text comprehension within videos, shares similarities with other studies dealing with text in natural images, commonly known as Scene Text VQA. The ST-VQA\cite{Biten_2019_ICCV} and TextVQA\cite{textvqa} datasets were the first to incorporate questions requiring understanding textual information from natural images. LoRRa\cite{textvqa} and M4C\cite{m4c} utilized pointer networks\cite{vinyals2015pointer} that generate answers from a fixed vocabulary and OCR tokens. In addition, M4C used a multimodal transformer\cite{vaswani2017attention} to integrate different modalities. TAP\cite{yang2021tap} employed a similar architecture to M4C and incorporated a pretraining task based on scene text, improving the model's alignment among the three modalities. Another study, LaTr\cite{biten2022latr}, focused on pretraining on text and layout information from document images and found that incorporating layout information from scanned documents improves the model's understanding of scene text.

\section{RoadTextVQA dataset}
This section looks at the data collection and annotation procedure, data analysis, and statistics.

\subsection{Data Collection}

The videos used in the dataset are taken from the RoadText-3K\cite{RoadText-3k} dataset and YouTube. The RoadText-3K dataset includes 3,000 ten-second road videos that are well-suited for annotation because they have a considerable quantity of text.
The RoadText-3K dataset includes videos recorded in the USA, Europe, and India and features text in various languages such as English, Spanish, Catalan, Telugu and Hindi. Each video contains an average of 31 tracks. However, the European subset is excluded from the annotation process for RoadTextVQA as it is dominated by texts in Spanish/Catalan, and the RoadTextVQA is   designed specifically  for  English road-text.
In addition to the videos from RoadText-3K, additional dashcam videos were sourced from the YouTube channel J Utah\footnote{ \url{https://www.youtube.com/@jutah}}. 252 videos from USA and UK were selected, and clips with a substantial amount of text were further selected by running a text detector over the video frames. Being a free and open-source text detector popular for scene text detection, we went with EasyOCR\cite{JaidedAI}  as our choice of text detector. The RoadText-3K videos have a resolution of 1280x720 with a frame rate of 30 frames per second. To keep the data consistent, the YouTube clips were downsampled to the same resolution and frame rate of 1280x720 at 30fps.\\

\begin{figure}[b]
\centering
  \includegraphics[width=0.9\linewidth]{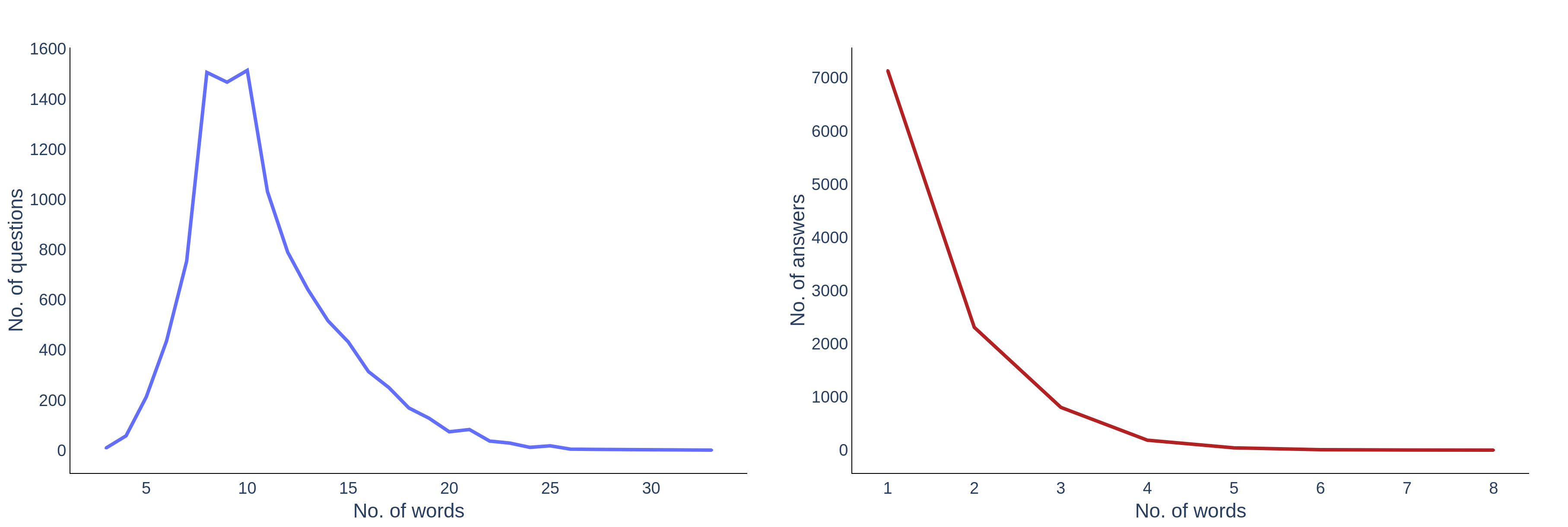}
  \caption{Distribution of the number of words in the question(left) and answer(right) of RoadTextVQA}\label{fig:roadtext_qn_ans}
\end{figure}

\begin{figure}[!htb]
\centering
\minipage{0.60\textwidth}
  \includegraphics[width=\linewidth]{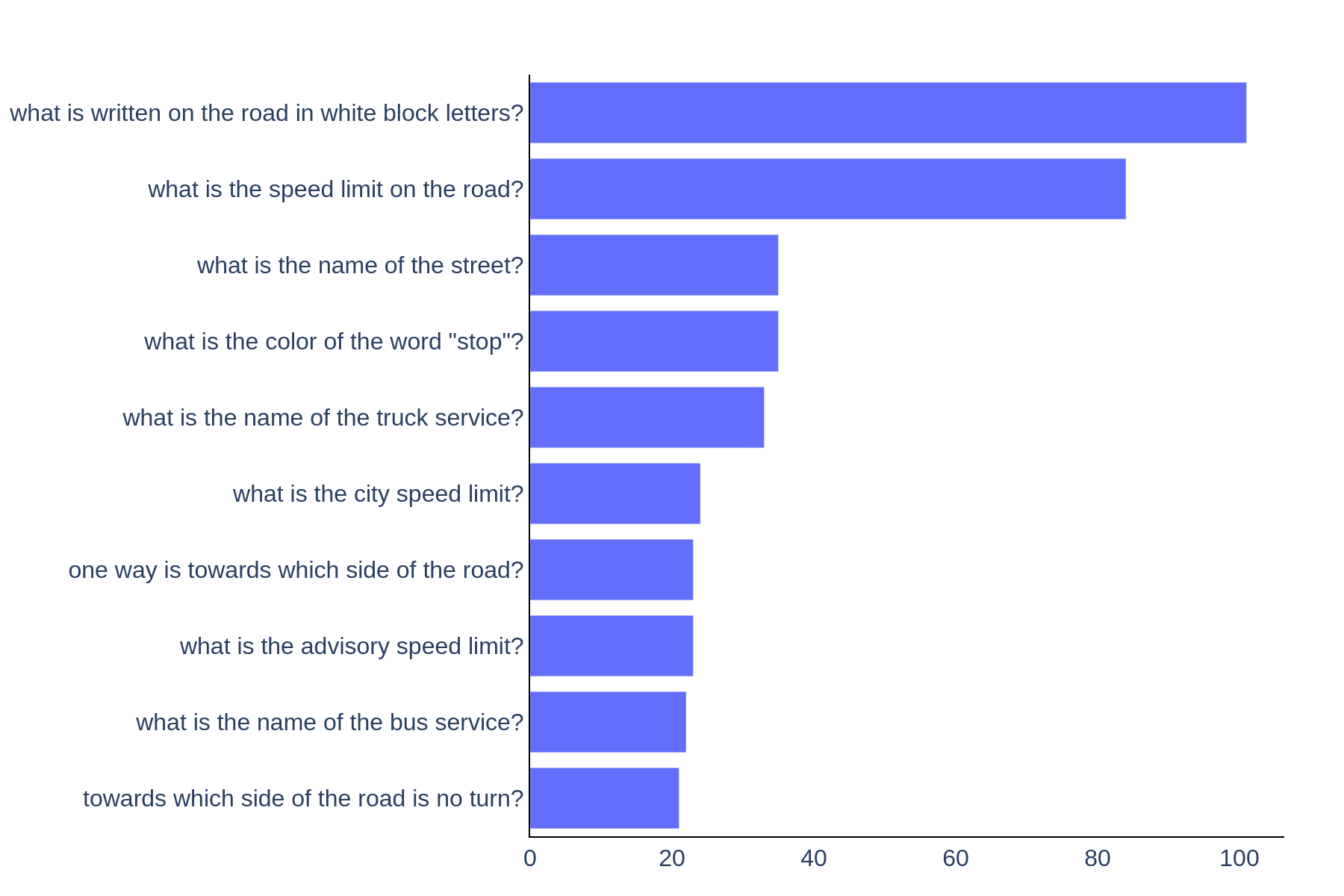}
  \caption{Top 10 questions in the dataset.}\label{fig:roadtext_qns}
\endminipage\hfill
\minipage{0.35\textwidth}
  \includegraphics[width=\linewidth]{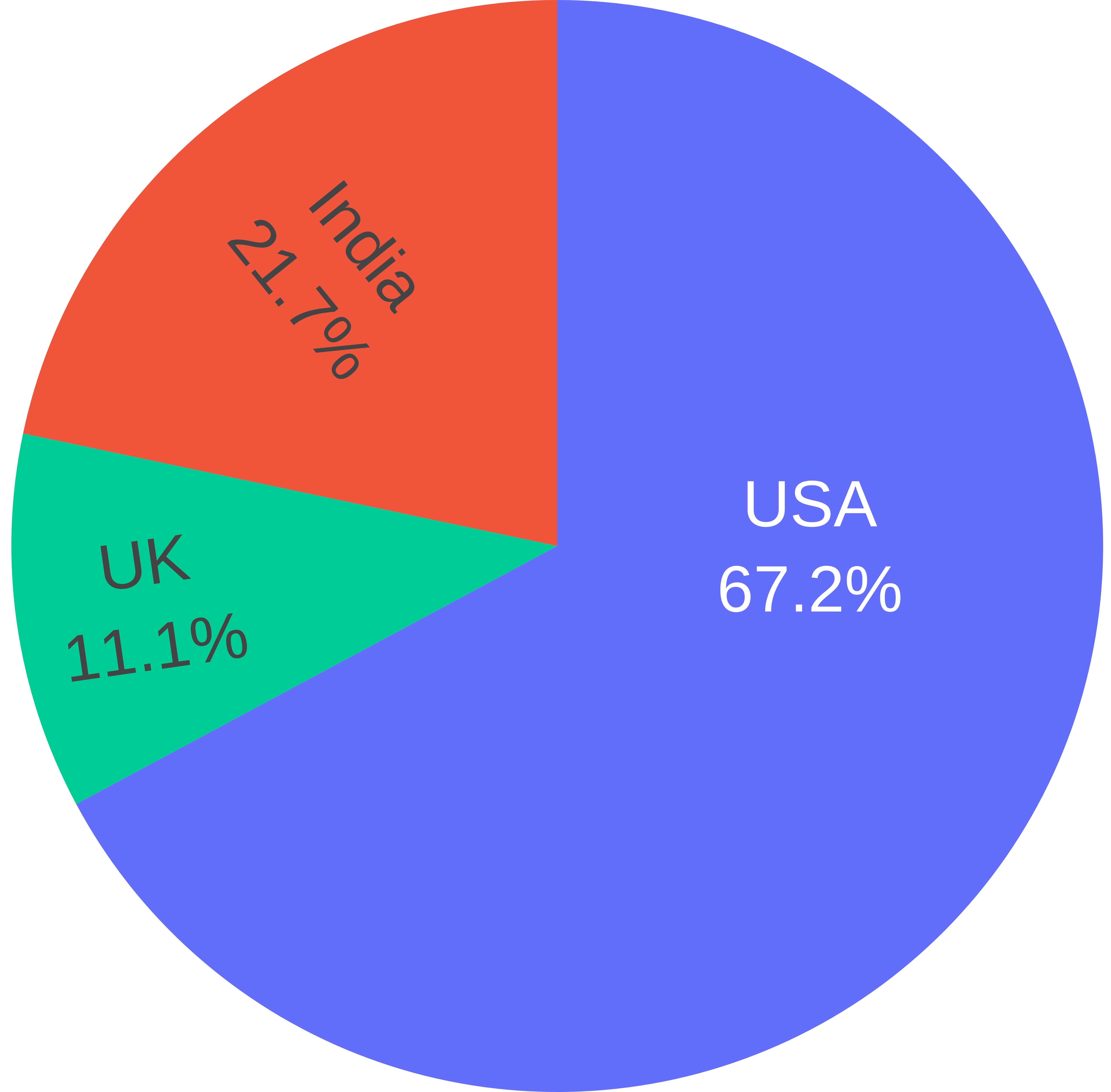}
  \caption{Geographical distribution of videos in the RoadTextVQA dataset.}\label{fig:roadtext_videos}
\endminipage
\end{figure}

Individuals who are proficient in the English language were hired to create the question-answer pairs. To ensure the quality of the applicants, an initial training session was conducted, followed by a filtering mechanism in the form of a comprehensive quiz. The quiz was designed to ensure that the question-answer pairs were created by individuals who had a solid grasp of the English language and a good understanding of the task, thereby enabling us to maintain a high standard of quality in the annotations.
The annotation process involved two stages, and a specifically designed web-based annotation tool was used. In the initial stage,  annotators add the question, answers and timestamp triads for videos shown to them. 
All the questions have to be based on either some text present in the video or on any road sign. In cases where a question could have multiple answers in a non-ambiguous way, the annotators were given the option to enter several answers. The timestamp is an additional data point which is collected, and it is the aptest point in the video at which the question is answerable. The annotators were instructed to limit the number of questions to not more than ten per video and to avoid asking any questions related to the vehicle license plate numbers. If there were no possible questions that could be asked from the video, then the annotators were given the option to reject it. 
In the verification stage, the video and the questions are shown, and the annotators had to add the answers and the timestamps. We made sure that verification is done by an annotator different from the one who has annotated it in the first stage.

\begin{table}[h]
    \centering
    \caption{Comparison of average and maximum question and answer lengths with other text based video question answering datasets.}
    \begin{tabular}{@{}lcccc@{}}
    \toprule
        Dataset & \multicolumn{2}{c}{  Average Length  } & \multicolumn{2}{c}{  Max Length  }\\
        \cmidrule(lr){2-3}
        \cmidrule(lr){4-5}
        & Question & Answer 
        & Question & Answer \\
    \midrule
        M4-ViteVQA\cite{zhao2022towards} & 6.75 & 1.94 & 24 & 26 \\
        NewsVideoQA\cite{newsvideoqa} & 7.04 & 2.02 & 20 & 19 \\
        RoadTextVQA & 10.78 & 1.45 & 33  & 8\\
    \bottomrule
    \end{tabular}
    \label{tab:avg_len_q_a}
\end{table}

\begin{figure}[!b]
    \centering
    \includegraphics[width=0.68\textwidth]{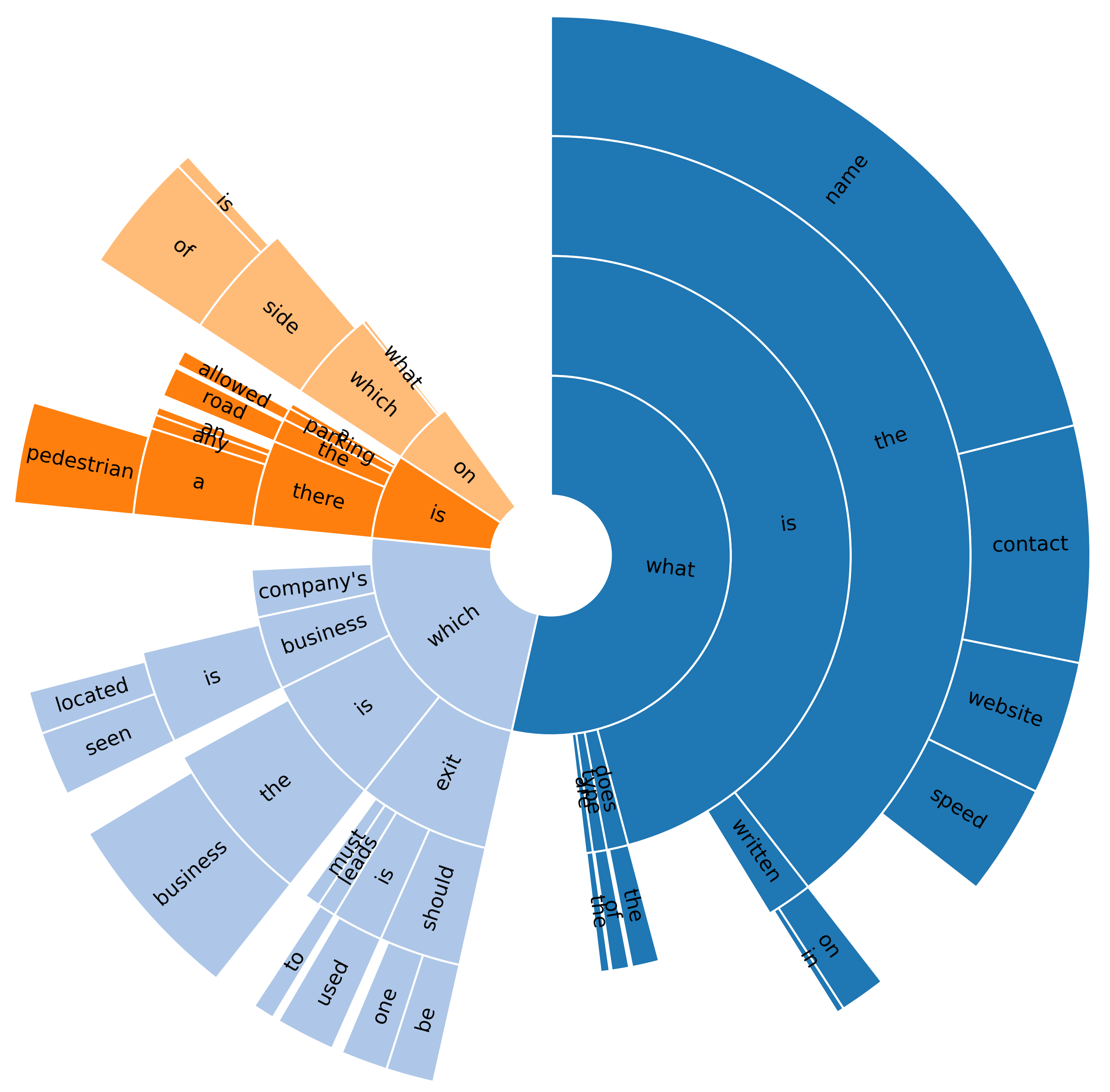}
    \caption{An analysis of the distribution of questions based on their starting 4 grams has shown that a significant proportion of questions are aimed at obtaining the name and contact information of businesses located along roads, as well as obtaining the speed limit for the road.} \label{roadtext_ngram}
\end{figure}

If the question is incorrect or does not follow the annotation guidelines, it is flagged and rejected. If for a question, there are common answers in the annotation stage and verification stage, then that question is considered valid. All the common answers are considered valid answers to the question.
In the verification stage, additional data regarding the question-answers are  also collected. The questions are categorically tagged into two distinct classes. Firstly, based on the type of question--- text-based or traffic sign-based.
The second classification captures whether the answer for a question, i.e., the text that makes up the answer, is present in the video or not.

\begin{figure}[t]
\centering
  \includegraphics[width=1\linewidth]{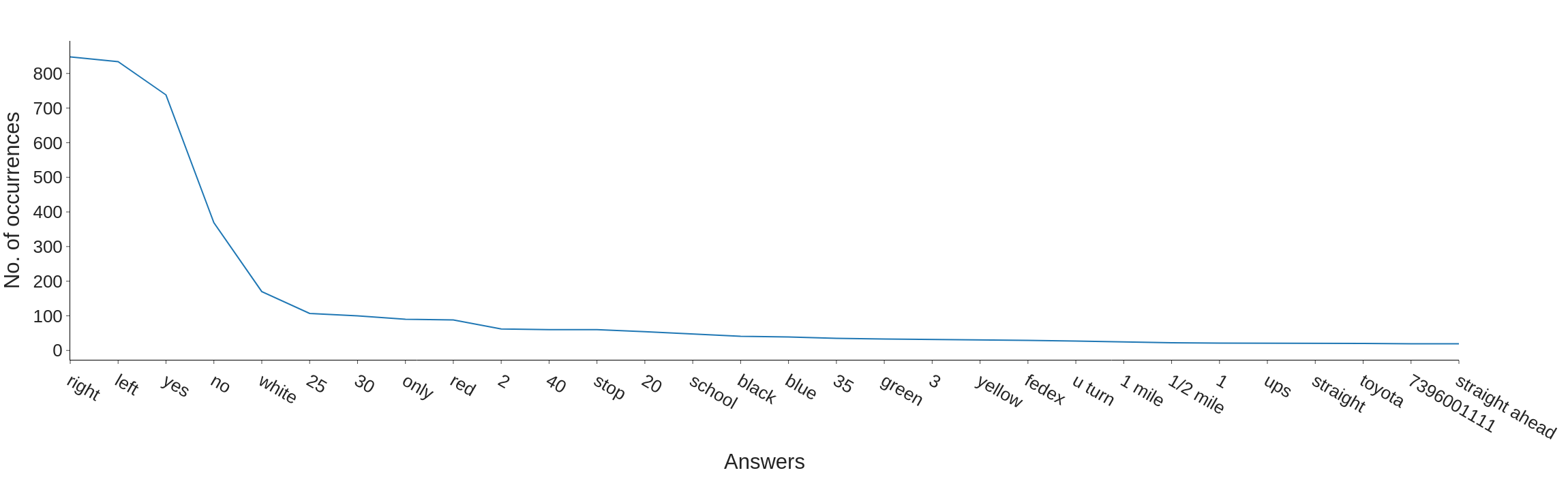}
  \caption{The number of occurrences of the answers in RoadTextVQA. The most recurring answer is "right", which makes up about 8\% of the answers.}\label{fig:roadtext_answers}
\end{figure}

\subsection{Data Statistics and Analysis}

The RoadTextVQA dataset contains 3,222 videos and 10,500 question-answer pairs. 
Among the  3,222 videos, 1,532 videos are taken from the RoadText-3K dataset and the rest are  from YouTube. 
The data is randomly split into 2,557 videos and 8,393 questions in the train set, 329 videos and 1,052 questions in the test, and 336 videos and 1,055 questions in the validation set.

The videos for the test and validation sets were randomly chosen from the RoadText-3K split, as it has ground truth annotations for text tracking. Methods that use OCR data can take advantage of the accurate annotations provided by RoadText-3K.

We present statistics related to the questions in RoadTextVQA through \autoref{fig:roadtext_qn_ans}, and \autoref{fig:roadtext_qns}. \autoref{fig:roadtext_qns} shows the most frequent questions and their frequencies. ``What is written on the road with white block letters?" is the most recurrent, followed by questions regarding the speed limits on the roads.

\begin{figure}[!t]
\minipage{\textwidth}
\centering
  \includegraphics[width=0.8\linewidth]{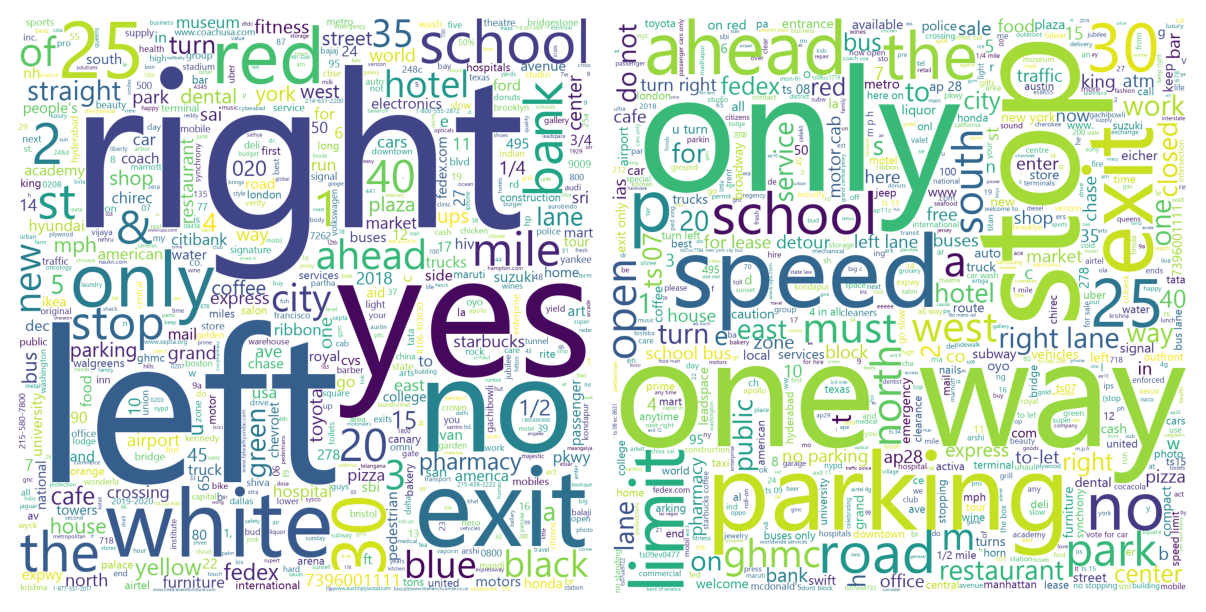}
  \caption{A visual representation of word frequency in the form of a word cloud, depicting the distribution of words in answers (left) and the distribution of OCR tokens from the videos (right).}\label{fig:wordcloud}
\endminipage\hfill
\end{figure}

\begin{figure}[!t]
    \centering
    \includegraphics[width=1\textwidth]{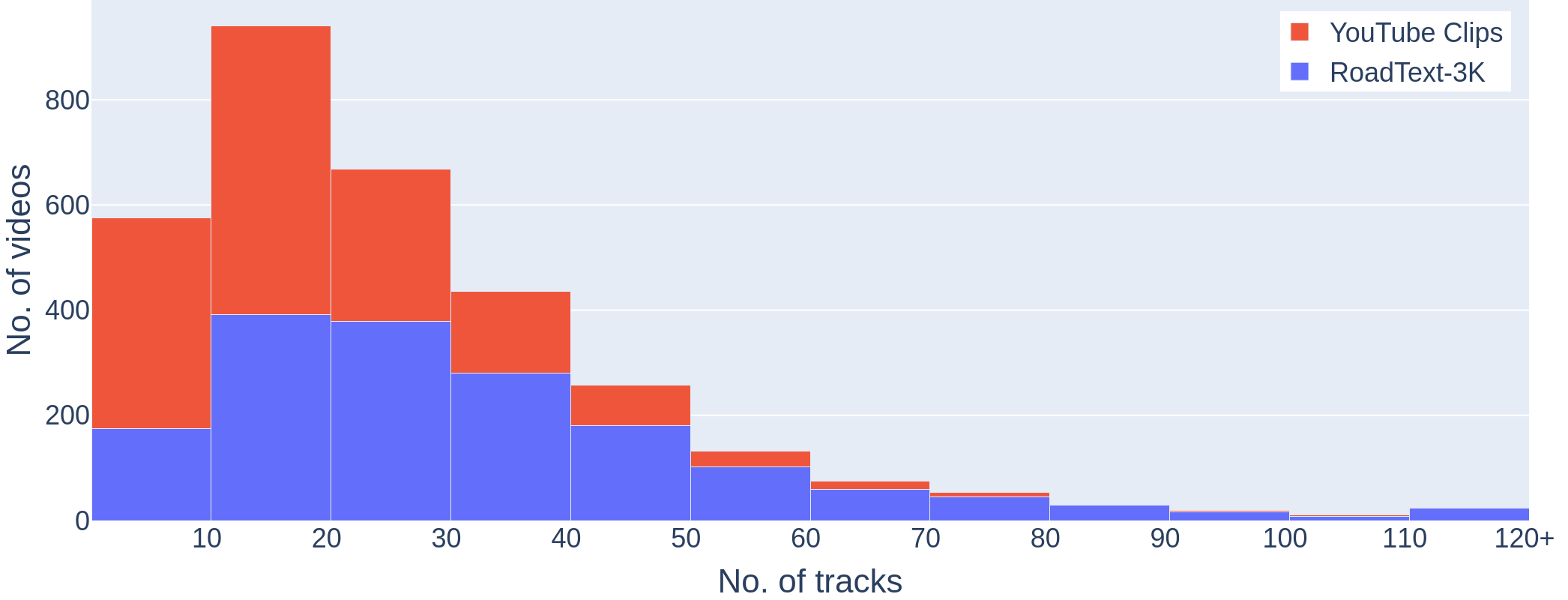}
  \caption{Distribution of number of videos vs number of tracks.}\label{fig:roadtext_ocr}
\end{figure}
\autoref{roadtext_ngram} provides a comprehensive overview of the question distribution in RoadTextVQA, with the majority of the questions being centred around details of shops located along the road. \autoref{fig:roadtext_qn_ans} depicts the word count in the questions and answers, respectively. The average number of words in the questions in RoadTextVQA is 10.8, while the average number in the answers is 1.45. The average number of words in questions is much higher when compared to other text-based VideoQA datasets, as seen in \autoref{tab:avg_len_q_a}. The percentage of unique questions stands at $86.6\%$, while the percentage of unique answers is $40.7\%$. \autoref{fig:roadtext_answers} shows the top 30 answers and the number of occurrences. \autoref{fig:wordcloud}, in the form of a word cloud, illustrates the most frequently occurring answers and OCR tokens. The most popular answers are ``right", ``left", ``yes", and ``no". The most prevalent OCR tokens in the videos are  ``stop", ``only", and ``one way".
The distribution of the videos in the dataset based on the geographic location where it was captured is shown in \autoref{fig:roadtext_videos}.
More than two-thirds  of the videos in the dataset are captured from roads in the USA. 

The majority of questions are grounded on text seen in the video ($61.8\%$), and the rest are based on road signs. Road signs can also contain text, such as speed limit signs or interchange exit signs. $68\%$ of questions have answers that can be found within the text present in the video, while the remaining $32\%$ of questions require an answer that is not a text  present in the video.

\section{Baselines}

This section presents details of the baselines we evaluate on the proposed RoadTextVQA dataset.
\subsection{Heuristic Baselines and Upper Bounds}

We evaluate several heuristic baselines and upper bounds on the dataset. These heuristics and upper bounds are similar to those used in other VQA benchmarks, such as TextVQA\cite{textvqa} and DocVQA\cite{docvqa_wacv}. The following heuristic baselines are  evaluated: 
(i) \textbf{Random Answer:} performance when answers to questions are randomly selected from the train split.
(ii) \textbf{Random OCR token:} performance when a random OCR token from the video is picked as the answer.
(iii) \textbf{Majority Answer:} performance when the most common answer in the train split is considered as the answer for all the questions.
The following upper bounds are evaluated
(i) \textbf{Vocab UB:} the upper bound on predicting the correct answer if it is present in the vocabulary of all the answers from the train split.
(ii) \textbf{OCR UB:} the upper bound on performance if the answer corresponds to an OCR token present in the video.
(iii) \textbf{Vocab UB + OCR UB}: this metric reflects the proportion of questions for which answers can be found in the vocabulary or the OCR transcriptions of the video.

\subsection{M4C}
The M4C\cite{m4c} model uses a transformer-based architecture to integrate representations of the image, question and OCR tokens. The question is embedded using a pretrained BERT\cite{devlin2018bert} model. Faster R-CNN\cite{ren2015faster} visual features are extracted for the objects detected and the OCR tokens in the image. 
The representation of an OCR token is formed from the FastText\cite{bojanowski2017enriching} vector, PHOC\cite{almazan2014word} vector, bounding box location feature, and Faster R-CNN feature of the token. A multi-head self-attention mechanism in transformers is employed, enabling all entities to interact with each other and model inter- and intra-modal relationships uniformly using the same set of transformer parameters. During answer prediction, the M4C model employs an iterative, auto-regressive decoder that predicts one word at a time. The decoder can use either a fixed vocabulary or the OCR tokens detected in the image to generate the answer. 

\subsection{SINGULARITY}
The architecture of SINGULARITY\cite{singularity} is made up of three major components: a vision encoder using ViT\cite{dosovitskiy2020vit}, a language encoder utilizing BERT\cite{devlin2018bert}, and a multi-modal encoder using a transformer encoder\cite{vaswani2017attention}. The multi-modal encoder uses cross-attention to collect information from visual representations using text as the key. Each video or image is paired with its corresponding caption during the pretraining phase, and the model is trained to align the vision and text representations using three losses (i) Vision-Text Contrastive: a contrastive loss which aligns the representations of vision and language encoders, (ii) Masked Language Modeling\cite{devlin2018bert}: masked tokens are predicted (iii) Vision-Text Matching: using the multi-modal encoder, predict the matching score of a vision-text pair. 
We use the SINGULARITY-temporal model, which is pretrained on 17M vision caption pairs\cite{cocotext,visual_genome,sbu,cc3m,cc12m,webvid}.
The SINGULARITY-temporal model contains a two-layer temporal encoder that feeds its outputs into the multi-modal encoder. SINGULARITY-temporal makes use of two new datasets named SSv2-Template Retrieval, and SSv2-Label Retrieval created from the action recognition dataset Something-Something v2 (SSv2)\cite{ssv2}. The pretraining is a video retrieval task using text queries. An additional multi-modal decoder is added for open-ended QA tasks and is initialised from the pretrained multi-modal encoder, which takes the multi-modal encoder's output as input and generates answer text with [CLS] as the start token.

\subsection{GenerativeImage2Text}
GIT(GenerativeImage2Text)\cite{wang2022git} is a transformer-based architecture aimed at unifying all vision-language tasks using a simple architecture pretrained on 0.8 billion image text pairs. GIT consists of an image encoder and a text decoder and is pretrained on a large dataset of image text pairs. The image encoder is a Swin-like\cite{liu2021Swin} transformer based on the contrastive pretrained model, which eliminates the need for other object detectors or OCR. As for the text decoder, the GIT uses a transformer with a self-attention and feed-forward layer to generate text output. The visual features and the text embeddings are concatenated and used as inputs to the decoder. GIT is able to gradually learn how to read the scene text with large-scale pretraining and hence achieves SoTA performance on scene-text-related VQA tasks such as ST-VQA. For video question answering, GIT employs a method of selecting multiple frames from the video and separately embeds each frame with a learnable temporal embedding which is initialized as zeros, and the image features are concatenated and used similarly to the image representation. The question and the correct answer are combined and used in the sense of a special caption, and the language model loss is computed solely on the answer and the [EOS] token.

\section{Experiments and Results}

This section covers the evaluation metrics, the experimental setup, and the experiment results.

\subsection{Experimental Setup}

\textbf{Evaluation metrics.} We use two evaluation metrics to evaluate the model's performance: Average Normalized Levenshtein Similarity (ANLS)\cite{Biten_2019_ICCV} and Accuracy (Acc. (\%)). The Accuracy metric calculates the percentage of questions where the predicted answer exactly matches any of the target answers.  
ANLS, on the other hand, does not award a zero score for all predictions that do not match the ground truth string exactly.
The score was originally proposed to act softly on cases where the predicted answer differs slightly from the actual. 
ANLS measures a similarity(based on the Levenshtein distance) between the prediction and ground truth and normalizes it as a  score in the range $[0,1]$. If the score is less than 0.5, the final ANLS score for the prediction is set to zero.

\textbf{OCR transcriptions.} The ground truth annotations were utilized for the videos in the RoadText-3K set, while for the remaining videos, the OCR transcriptions were sourced using the Google Cloud Video Intelligence API. Both RoadText-3K ground truth annotations, and the Google API provide text transcriptions at the line level. 
We use the line-level text transcriptions as the OCR tokens for the calculation of OCR upper bounds and OCR-based heuristics as given in the \autoref{tab:roadtext_hu}. When a text track gets cut off from the frame or partially occluded by other objects in a video, the Google Cloud Video Intelligence API treats it as a new track, whereas RoadText-3K annotations ignore the partially occluded tracks. This is why in the \autoref{fig:roadtext_ocr}, the number of videos vs the number of tracks is a bit inflated for the YouTube clips when compared to RoadText-3K clips. 

\textbf{Experimental setup for M4C.} 
The M4C\cite{m4c} model is trained using the official implementation, and the training parameters and implementation details remain consistent with those used in the original paper. We used a fixed vocabulary of size 3926 generated from the train set. 
The training data consists of image question-answer pairs where the image selected for training is the one on which the questions are based, specifically the timestamp frame. After training, the model is evaluated using two approaches. Firstly, it is tested on the timestamp QA pairs of the test set, and secondly, it is evaluated on the video level by sampling ten frames from the respective video for each QA pair and obtaining the model prediction for every frame individually. The final answer is determined by taking the most common answer from the ten individual frame predictions.
\\

\textbf{Experimental setup for SINGULARITY.} 
We fine-tuned the pretrained SINGULARITY-temporal 17M model on four NVIDIA Geforce RTX 2080 Ti. The fine-tuning process was run for 20 epochs with a batch size of 16, starting with an initial learning rate of 1e-5 and increasing linearly in the first half epoch, followed by cosine decay\cite{sdgr} to 1e-6. The other parameters used for training are the same as the official implementation. The video frames were resized to 224x224, and a single frame with random resize, crop and flip augmentations was utilised during training, whereas 12 frames were used during testing. Additionally, we fine-tuned the SINGULARITY model, which has been pretrained on the MSRVTT-QA\cite{xu2017video} dataset.\\

\textbf{Experimental setup for GIT.}
The training process for GIT was carried out using a single Tesla T4 GPU for 20 epochs with a batch size of 2.
We use an Adam\cite{adam} optimizer with an initial learning rate starting at 1e-5 and gradually decreasing to 1e-6 through the use of cosine decay.
The GIT model was trained using the official VideoQA configuration used for MSRVTT-QA training. We fine-tuned the pretrained GIT-large model on our dataset, using six frames that were evenly spaced as inputs during both training and testing. In addition, we further fine-tuned the GIT model that was pretrained on the MSRVTT-QA\cite{xu2017video} dataset.

\subsection{Results}
Heuristic baselines and upper bound results are presented in the \autoref{tab:roadtext_hu}. The heuristic baselines yield very low accuracy, which indicates the absence of any bias due to the repetition of answers. 

\begin{table}
    \caption{Performance of various heuristic baselines and upper bounds that are commonly evaluated on text-based VQA datasets.}
    \centering
    \begin{tabular}{@{}lcc@{}}
    \toprule
        Baseline & \multicolumn{2}{c}{Test} \\
        \cmidrule(lr){2-3}
        & ANLS & Acc.(\%) \\
    \midrule
        Random Answer & 0.09 & 0 \\
        Random OCR token & 3.20 & 1.98 \\
        Majority Answer & - & 3.49 \\
        Vocab UB & - & 59.26 \\
        OCR UB & - & 36.67 \\
        Vocab + OCR UB & - & 76.18 \\
    \bottomrule
    \end{tabular}
    \label{tab:roadtext_hu}
\end{table}

\begin{table*}[!b]
    \caption{Performance of RoadTextVQA on M4C. Abbreviations- TB: text-based questions, RSB: road sign-based questions, AP: questions where the answer is present in the video, ANP: questions where the answer is not present in the video.}
    \centering
    \begin{adjustbox}{width=1\textwidth}
    \begin{tabular}{@{}lcccccccccc@{}}
    \toprule
        Test Frames & \multicolumn{2}{c}{TB} & \multicolumn{2}{c}{RSB} & \multicolumn{2}{c}{AP} & \multicolumn{2}{c}{ANP} & \multicolumn{2}{c}{All} \\
        \cmidrule(r){2-3} \cmidrule(r){4-5}\cmidrule(l){6-7}\cmidrule(l){8-9}\cmidrule(lr){10-11}
         & ANLS & Acc. (\%) & ANLS & Acc. (\%) & ANLS & Acc. (\%) & ANLS & Acc. (\%)  & ANLS & Acc. (\%) \\
    \midrule
        1 & 35.28 & 29.27 & 55.49 & 49.46 & 37.55 & 29.70 & 63.85 & 63.19 & 44.22 & 38.20 \\
        10 & 23.92 & 21.48 & 42.83 & 38.32 & 20.38 & 15.96 & 67.12 & 66.91 & 32.27 & 28.92 \\
    \bottomrule
    \end{tabular}
    \end{adjustbox}
    \label{tab:roadtext_m4c}
\end{table*}

Random OCR heuristic gives close to 2\% accuracy, meaning that there is enough text present in the video that selecting a random OCR from the video will not yield high accuracy. The OCR upper bound is 36.6\% which is low when compared to the percentage of questions which have the answers present in the video. The low OCR UB can be attributed to how the text detection and how ground truth annotation is done. The response to a question may be split into multiple lines within the video, leading to the representation of the answer as separate tokens in the OCR output. This happens because the annotations in the OCR process were carried out on a line level. From the upper bound result of Vocab  + OCR UB, we can see that more than three-quarters of the answers are present in either the vocabulary or in the OCR tokens of the video.
\begin{table*}[!tpbh]
    \caption{Performance of RoadTextVQA on SINGULARITY and GIT. Abbreviations- TB: text-based questions, RSB: road sign-based questions, AP: questions where the answer is present in the video, ANP: questions where the answer is not present in the video.}
    \centering
    \begin{adjustbox}{width=1\textwidth}
    \begin{tabular}{@{}lccccccccccc@{}}
    \toprule
        Method & Pretrain Data & \multicolumn{2}{c}{TB} & \multicolumn{2}{c}{RSB} & \multicolumn{2}{c}{AP} & \multicolumn{2}{c}{ANP} & \multicolumn{2}{c}{All} \\
        \cmidrule(lr){3-4}\cmidrule(r){5-6} \cmidrule(r){7-8}\cmidrule(l){9-10}\cmidrule(l){11-12}
        & & ANLS & Acc. (\%) & ANLS & Acc. (\%) & ANLS & Acc. (\%) & ANLS & Acc. (\%)  & ANLS & Acc. (\%) \\
    \midrule
        SINGULARITY & - & 15.38 & 14.04 & 45.29 & 33.04 & 17.36 & 9.22 & 61.71 & 61.33 & 28.62 & 22.45 \\
        SINGULARITY & MSVRTT-QA & 17.25  & 15.22 & 47.84 & 36.46 & 19.50 & 11.50 & 63.98 & 63.19 & 30.79 & 24.62 \\
        \hline
        GIT & - & 18.09 & 14.38 & 50.36 & 39.82 & 20.98 & 12.16 & 65.65 & 65.05 & 32.34 & 25.61 \\
        \textbf{GIT} & MSRVTT-QA &   \textbf{22.61} & \textbf{19.62} & \textbf{51.20} & \textbf{42.18} & \textbf{23.40} & \textbf{15.96} & \textbf{69.93} & \textbf{69.51} & \textbf{35.23} & \textbf{29.58} \\
    \bottomrule
    \end{tabular}
    \end{adjustbox}
    \label{tab:roadtext_sing}
\end{table*}

\begin{figure}[t]
\centering
  \includegraphics[width=1\linewidth]{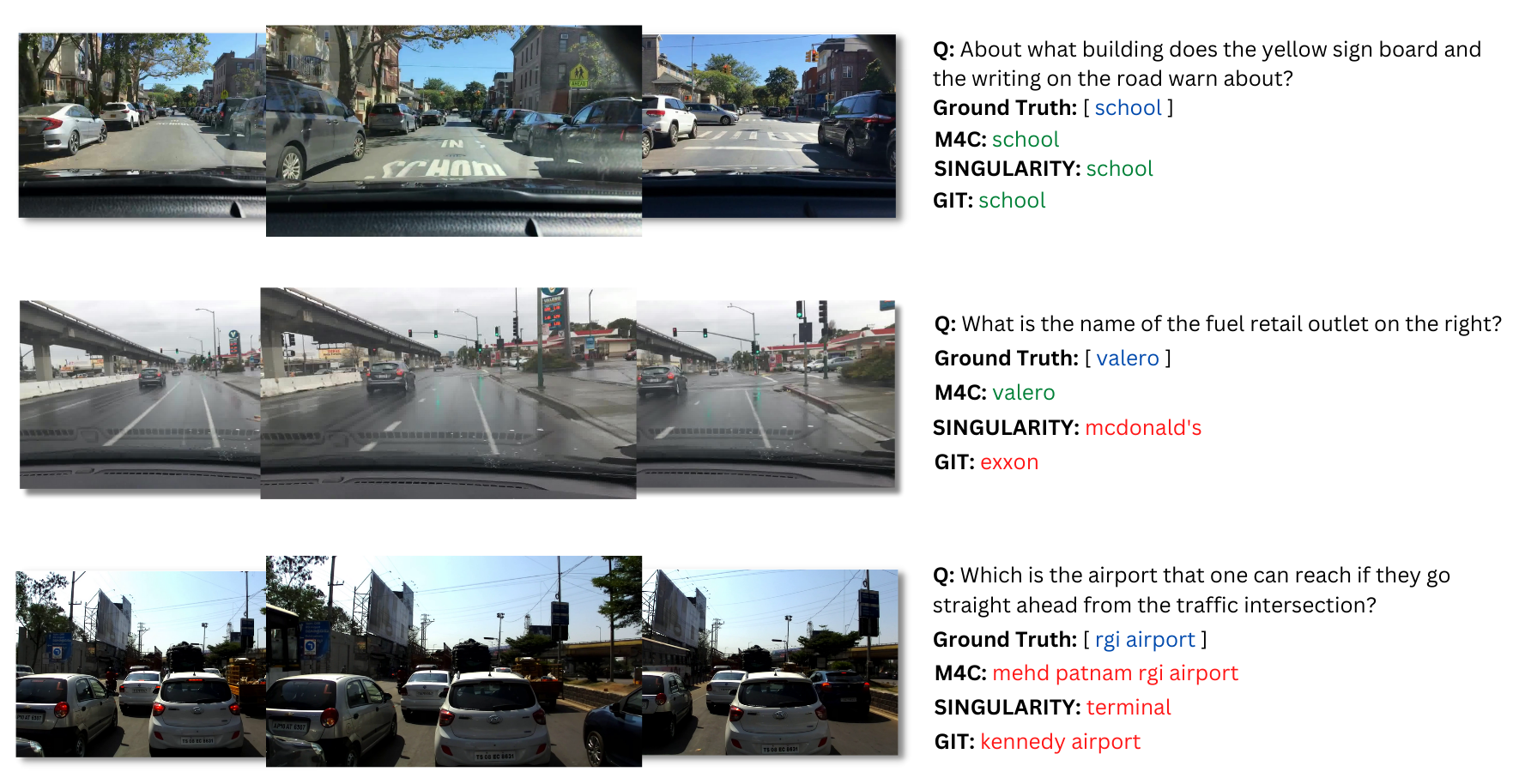}
  \caption{Qualitative results showing predictions of M4C, SINGULARITY and GIT. The correct predictions are highlighted in green, whereas the incorrect ones are highlighted in red.}\label{fig:roadtext_qual}
\end{figure}

The results on  M4C are shown on \autoref{tab:roadtext_m4c}. The frame level results, where we evaluate on the timestamp frame, show an accuracy of 38.20\% and the video level results, where we evaluate on ten frames, give an accuracy of 28.92\%. The results show that answering the question is still a challenging task, even when we reduce the complexity of the problem by providing the aptest frame for answering the question and ground truth OCR tokens. \\
We show the results after fine-tuning on SINGULARITY and GIT in \autoref{tab:roadtext_sing}. The accuracy of the questions requiring answers to be extracted from the video (AP) is comparatively lower, while the accuracy of the questions where the answer is not present in the video is comparatively higher.
Compared to AP, ANP is less complex to answer because it involves a fixed set of answers. In contrast, AP requires dynamic extraction from OCR tokens, resulting in the ANP set having better accuracy than AP.
Additionally, fine-tuning the model that has been pretrained on the MSRVTT-QA dataset shows improvement in accuracy across all categories(TB, RSB, AP, and ANP). 

Fine-tuning GIT results in better performance compared to SINGULARITY. GIT also shows a similar trend when fine-tuned on pretrained MSRVTT-QA dataset. The ``answer is present in the videos(AP)" subset has an improvement of 3.9\% in accuracy when compared with  SINGULARITY, whereas the ``answer is not present(ANP)" in the videos subset has a gain of 6.3\%. M4C tested on a single frame shows better results compared to VideoQA models. This can be attributed to the fact that we explicitly provide the OCR tokens and the correct frame on which the question is framed to the model. M4C tested on ten frames gives comparable results to GIT. 

We show some of the qualitative results in \autoref{fig:roadtext_qual}. As the complexity of the scene and the obscurity of the scene text increase, it becomes more and more difficult for the model to predict the correct answer. VideoQA baselines achieve better results on questions that do not require the extraction of answers from the video.

\section{Conclusions}
We introduce RoadTextVQA, a new Video Question Answering dataset where the questions are grounded on the text and road signs present in the road videos. Our findings from the baseline models' performance indicate a need for improvement in existing VideoQA approaches for text-aware multimodal question answering. 

Future work can involve augmenting the dataset by incorporating videos obtained from diverse global locales. Currently, there are recurrent questions and answers due to repeating elements in the videos.
Including videos from various locations broadens the diversity of the dataset by providing a more comprehensive range of questions and answers and minimizes any biases within the dataset. To our best knowledge, currently, there are no Visual Question Answering models that explicitly incorporate road signs. Models can integrate road signs as an additional input or pretrain on road sign-description pairs to enhance their ability to respond to questions that require domain knowledge. 

We believe this work would encourage researchers to develop better models that incorporate scene text and road signs and are resilient to the challenges posed by driving videos. Additionally, drive further research in the area of scene text VideoQA and the development of advanced in-vehicle support systems.

\section{Acknowledgements}
This work has been supported by IHub-Data at IIIT-Hyderabad, and grants PDC2021-121512-I00, and PID2020-116298GB-I00 funded by MCIN/AEI/ \\10.13039/501100011033 and the European Union NextGenerationEU/PRTR.

%
%
%
\bibliographystyle{splncs04}
\bibliography{main}

\end{document}